# Improved lightweight identification of agricultural diseases based on MobileNetV3

Yuhang Jiang[1,*], Wenping Tong[2]
[1]School of Internet Anhui University, Hefei, China
[2]School of Internet Anhui University, Hefei, China
[*]yuhang.tjtj@foxmail.com

## Abstract

At present, the identification of agricultural pests and diseases has the problem that the model is not lightweight enough and difficult to apply. Based on MobileNetV3, this paper introduces the Coordinate Attention block. The parameters of MobileNetV3-large are reduced by 22%, the model size is reduced by 19.7%, and the accuracy is improved by 0.92%. The parameters of MobileNetV3-small are reduced by 23.4%, the model size is reduced by 18.3%, and the accuracy is increased by 0.40%. In addition, the improved MobileNetV3-small was migrated to Jetson Nano for testing. The accuracy increased by 2.48% to 98.31%, and the inference speed increased by 7.5%. It provides a reference for deploying the agricultural pest identification model to embedded devices.

## 1    Introduction

Crop diseases and insect pests are one of the important disasters restricting agricultural production, seriously affecting the yield and quality of crops. In 2020, the cumulative disaster area in China reached 300 million hectares. In recent years, China has reduced food losses by 87 to 110 million tons each year by adopting various pest and disease prevention and control measures, accounting for 16.00% to 19.55% of the country's total grain. It can be seen that the monitoring of crop diseases and insect pests plays a very important role in agricultural production [1].

In recent years, with the rapid development of the field of deep learning, many models have also achieved good results in the field of crop pest identification [2]. Such as GoogLeNet [6] is widely used. However, many models are difficult to apply in practice due to the large amount of parameters.

To solve the problem of model deployment, researchers have proposed many lightweight neural network architectures. MobileNet [3] uses depthwise separable convolution to reduce the model size and the number of parameters. EfficientNet [7] uses compound scaling, which can scale depth, width, resolution evenly by a uniform coefficient. EfficientNetV2 [8] introduces a powerful training trick: Progressive learning. It can reduce training time. They provide strong support for the practical application of crop pest and disease identification. With the popularity of BERT and Transformer, the attention mechanism has also affected the field of computer vision.

The attention mechanism is a special structure embedded in the machine learning model, which is used to automatically learn and calculate the contribution of the input data to the output data. This mechanism is effective. For example, an improvement of MobileNetV3 is the addition of SE block [9].

Although there are many excellent network architectures, a lot of research work is still in the laboratory. Many deep learning related research work has also begun to be carried out on embedded devices [12-14]. Agricultural machinery is a field that pays great attention to practical applications. Therefore, this paper proposes a lightweight crop pest identification method based on MobileNetV3. We introduce the Coordinate Attention [10] mechanism and test on the PlantVillage [11] dataset. Finally, we transfer the trained model to run on NVIDIA Jetson Nano. Compared with the original MobileNetV3, the model is smaller and more accurate. On embedded devices, the MobileNetV3-small+CA proposed in this paper improves the accuracy by 2.48% to 98.31%, and increases the inference speed by 7.5%.

## 2    Related work

### 2.1  Disadvantages of MobileNetV3

MobileNetV3 [5] is a lightweight network combining MobileNetV1 [3] and MobileNetV2 [4], which has higher accuracy and efficiency. Based on the structure of MobileNetV2, MobileNetV3 introduces the SE (Squeeze-and-Excitation) attention module. SE attention module effectively builds the interdependencies between channels by simply squeezing each 2D feature map.

The SE module main includes two part: Squeeze and Excitation. After completing the above two steps, 'Scale' is used to multiply the channel weights. The specific operation is that the SE module calculates the weight value of each channel. Then SE module multiplies the weight values with the two-dimensional matrix of the corresponding channel of the original feature map.

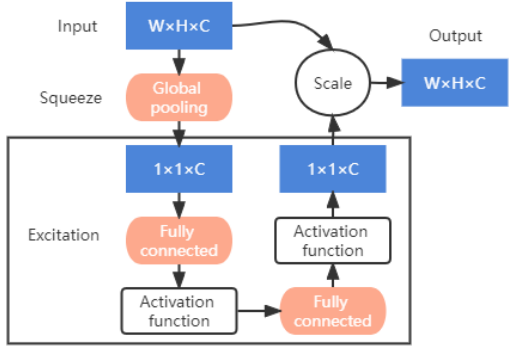

**Figure 1** SE module workflow.

However, it only considers re-weighting the importance of each channel by modeling channel relationships, ignoring the location information. The location information is important for generating spatially selective attention maps. Moreover, the SE module will increase the total number of parameters and the total amount of calculation of the network. Although the amount of calculation of the fully connected layer used is not larger than that of the convolutional layer, the amount of parameters will increase significantly. So we replaced the SE block with the Coordinate Attention Block to improve the network.

## 2.2 Coordinate Attention Block

The Coordinate Attention (CA) considers a more efficient way to capture location information and channel relationships to enhance the feature representation of Mobile Networks. The specific operation of CA is divided into two steps: Coordinate information embedding and Coordinate Attention generation. It is a channel attention + x direction space + y direction space attention block.

### 2.2.1 Coordinate information Embedding

Coordinate Attention decomposed the global pooling and converted into one-to-one 1D feature encoding operation, the formula is summarized:

$$z_c = \frac{1}{H \times W} \sum_{i=1}^{H} \sum_{j=1}^{W} x_c(i,j) \qquad (1)$$

where $z_c$ is the output associated with the c-th channel.

Then, Each channel is first encoded along the horizontal and vertical coordinates using a pooling kernel of size (H, 1) or (1, W), respectively. The output of the c-th channel at height j can be expressed as follows:

$$z_c^h(h) = \frac{1}{W} \sum_{0 \le i < W} x_c(h,i) \qquad (2)$$

Similarly，the output of the c-th channel at weight j can be expressed as follows:

$$z_c^w(h) = \frac{1}{H} \sum_{0 \le j < H} x_c(j,w) \qquad (3)$$

The above two transformations can extract features along two directions respectively, and obtain a pair of feature maps based on direction perception. Better than SE blocks that generate a single feature vector. The CA block helps the network locate more interesting targets.

### 2.2.2 Coordinate information Generation

Coordinate information generate is design for better use of features generated by coordinate information embedding. Howard, Andrew, et al [5] mainly refer to the following 3 standards:
First, the new transformation should be as simple as possible for applications in the Mobile environment;
Second, it can make full use of the captured location information, so that the region of interest can be accurately captured;
Finally, it should also be able to efficiently capture the relationship between channels.
After a series of formula transformations, the coordinate information generation can be summarized:

$$y_c(i,j) = x_c(i,j) \times g_c^h(i) \times g_c^w(j) \qquad (4)$$

where $g_c$ is the result of channel number transformation and convolution transformation.

### 2.2.3 Advantages of Coordinate Attention

The Coordinate Attention mechanism enables efficient positioning on the pixel coordinate system. It enables the model to focus on the area of interest and obtain information in a larger area, so as to achieve better classification results.

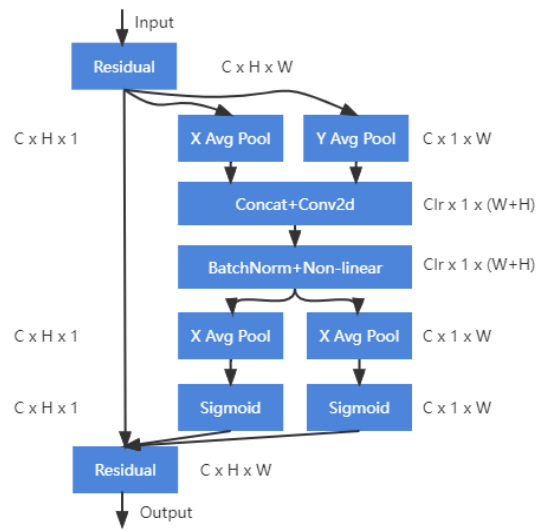

**Figure 2** CA block workflow, 'X Avg Pool' and 'Y Avg Pool' refer to 1D horizontal global pooling and 1D vertical global pooling.

## 2.3 Network structure improvement

The SE block only considers the information encoding between channels, but ignores the spatial information. This wastes the information obtained by bneck's $5\times5$ convolution kernel in MobileNetV3. So we replace the corresponding SE block with CA block. This not only captures long-range correlations along one direction, but preserves precise location information along the other. Moreover, due to the lower calculation amount of the CA block, the computational burden brought by the $5\times5$ convolution kernel can be offset. See table 1 and 2 for details.

**Table1** Specification for bnecks in MobileNetV3-large+CA. In the column of attention block, '0' means not using the attention block, '1' means using SE block, '2' means using CA block.

| bneck id | Kernel Size | exp size | #out | Attention block |
|---|---|---|---|---|
| 1 | $3\times3$ | 16 | 16 | 0 |
| 2 | $3\times3$ | 64 | 24 | 0 |
| 3 | $3\times3$ | 72 | 24 | 0 |
| 4 | $5\times5$ | 72 | 40 | 2 |
| 5 | $5\times5$ | 120 | 40 | 2 |
| 6 | $5\times5$ | 120 | 40 | 2 |
| 7 | $3\times3$ | 240 | 80 | 0 |
| 8 | $3\times3$ | 200 | 80 | 0 |
| 9 | $3\times3$ | 184 | 80 | 0 |
| 10 | $3\times3$ | 184 | 80 | 0 |
| 11 | $3\times3$ | 480 | 112 | 1 |
| 12 | $3\times3$ | 672 | 112 | 1 |
| 13 | $5\times5$ | 672 | 160 | 2 |
| 14 | $5\times5$ | 960 | 160 | 2 |
| 15 | $5\times5$ | 960 | 960 | 1 |

**Table2** Specification for bnecks in MobileNetV3-large+CA. See table 1 for notation.

| bneck id | Kernel Size | exp size | #out | Attention block |
|---|---|---|---|---|
| 1 | $3\times3$ | 16 | 16 | 1 |
| 2 | $3\times3$ | 72 | 24 | 0 |
| 3 | $3\times3$ | 88 | 24 | 0 |
| 4 | $5\times5$ | 96 | 40 | 2 |
| 5 | $5\times5$ | 240 | 40 | 2 |
| 6 | $5\times5$ | 240 | 40 | 2 |
| 7 | $5\times5$ | 120 | 48 | 2 |
| 8 | $5\times5$ | 144 | 48 | 2 |
| 9 | $5\times5$ | 288 | 96 | 2 |
| 10 | $5\times5$ | 576 | 96 | 2 |

## 2.4 Migration verification

For the embedded device on which the model was deployed, we chose the Jetson Nano produced by NVIDIA. The Jetson Nano is a small AI computer with decent performance and power consumption at an affordable price. It can run modern AI workloads, run multiple neural networks in parallel, and process data from multiple high-resolution sensors simultaneously. This makes it an ideal entry-level option for adding advanced AI to embedded products. Its part of technical specifications are showed in Table3. And Figure 3 is our Jetson Nano board.

**Table 3** A part of technical specifications of Jetson Nano.

| Parameter | Technical specifications |
|---|---|
| CPU | Quad-core ARM® Cortex®-A57 MPCore processor |
| GPU | NVIDIA Maxwell™ architecture with 128 NVIDIA CUDA® cores 0.5TFLOPS (FP16) |
| Memory | 4 GB 64-bit LPDDR4 1600 MHz – 25.6 GB/s |
| Storage | 16 GB eMMC 5.1 Flash |
| Video decoding | 500 MP/s 1x 4K @ 60 (HEVC) 2x 4K @ 30 (HEVC) 4x 1080p @ 60 (HEVC) 8x 1080p @ 30 (HEVC) |
| Camera | 2 lanes (3x4 or 4x2) MIPI CSI-2 D-PHY 1.1 (18 Gbps) |
| Size | 69.6mm x 45mm |
| Prize | $99 |

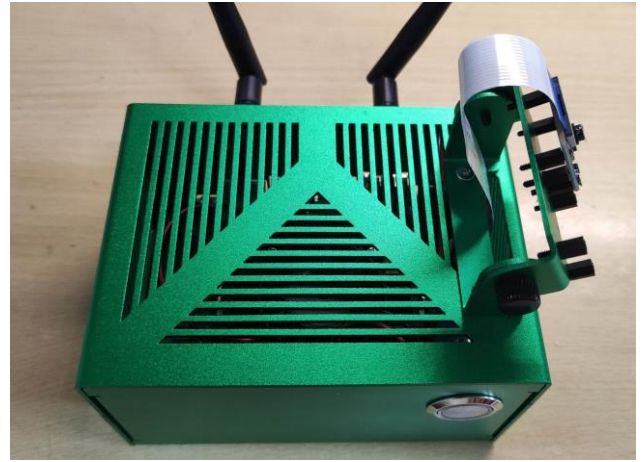

**Figure 3** NVIDIA Jetson Nano

## 3 Model training

### 3.1 Datasets and Preprocessing

Dataset We use the public dataset PlantVillage [11] on Kaggle. It contains 54305 images, each image is $256\times256$ pixels. We get pictures of 13 types of crops such as apples, tomatoes, strawberries, potatoes, etc., a total of 38 categories (including health and disease pictures). Figure 3 is the picture data of healthy apple's leaf and apple's leaf with scab.

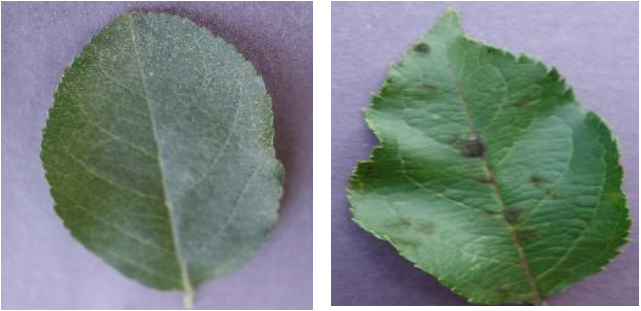

**Figure 4** Left is a healthy apple leaf, right is a diseased apple leaf.

On the basis of the original image, we make random horizontal offset, vertical offset, and horizontal flip to achieve the effect of data enhancement. At the same time, it also simulates data collection scenarios in real life to enhance the generalization of the model. After image augmentation, we divide the training set, validation set and test set in a ratio of 7:1:2. Finally convert the image to 224x224 pixels and input it into the model for training.

### 3.2 Experimental details

The experiments in this paper are carried out on workstations and embedded systems, respectively. We compared with the current advanced classic lightweight networks.

Workstation environment: AMD EPYC 7642 48-Core Processor, NVIDIA RTX 3090, Ubuntu 20.04 operating system, Tensorflow 2.3.
Compare models: MobileNetV3-small, MobileNetV3-large, shufflenet_v2_x0_5, shufflenet_v2_x1_0, shufflenet_v2_x2_0, GoogLeNet.

Jetson Nano operating environment: Ubuntu 18.04 operating system, Tensorflow 2.3, opencv 4.1.1, jetpack 4.4.1.
Compare models: MobileNetV3-small, MobileNetV3-large, shufflenet_v2_x0_5, shufflenet_v2_x1_0, shufflenet_v2_x2_0.

Each model is trained for 30 epochs, the optimizer is Adam, the learning rate is 0.001, the exponential decay rate of the first moment estimation is 0.9, the exponential decay rate of the second moment estimation is 0.9, and the epsilon value is 1e-8. Finally, we calculate the accuracy, model parameters, FLOPs, and model size.

## 4 Results and Analysis

As shown in Table 4, MobileNetV3-large and MobileNetV3-small added CA block, the performance is improved, and the model is more lightweight, which can be said to serve multiple purposes.

**Table 4** Comparisons of the performance of different models. Underlined parameters represent the best parameters for that column

| Model | Accuracy (%) | Params | FLOPs (G) | Size (M) |
| --- | --- | --- | --- | --- |
| shufflenet_v2_x0_5 | 94.79 | 388,694 | <u>0.082</u> | <u>10.3</u> |
| shufflenet_v2_x1_0 | 96.09 | 1,308,734 | 0.293 | 20.0 |
| shufflenet_v2_x2_0 | <u>98.44</u> | 5,456,574 | 1.17 | 68.1 |
| MobileNetV3-large | 97.39 | 4,275,110 | 0.446 | 54.3 |
| MobileNetV3-small | 96.35 | 1,568,918 | 0.117 | 22.4 |
| GoogLeNet | 94.14 | 6,012,502 | 0.05 | 25.2 |
| **MobileNetV3-small+CA (ours)** | 96.74 | <u>1,202,347</u> | 0.119 | 18.3 |
| **MobileNetV3-large+CA (ours)** | 98.31 | 3,333,799 | 0.449 | 43.6 |

It can be seen from the comparison that the introduction of CA block has brought a very significant and excellent effect. The experimental results show that the CA block improves the accuracy of MobileNetV3-large by 0.92%, reduces the parameters by 22.0%, and reduces the model size by 19.7%, while only adding a few FLOPs to MobileNetV3. MobileNetV3-small Params decreased by 23.4%, model size decreased by 18.3%, and accuracy increased by 0.40%.

In the application of lightweight networks, it is not only necessary to look at the level of a certain indicator, but to comprehensively weigh each indicator. Although shufflenet_v2_x2_0 has the highest accuracy, its parameters, FLOPs, and size are all the largest, which cannot meet the application requirements. shufflenet_v2_x0_5 has the smallest size and FLOPs but the second-to-last accuracy, only 94.79%.

On the whole, MobileNet+CA with low model parameters and computational consumption and high test accuracy has high cost performance.

However, the operating efficiency of the model will be affected by the performance of the computing platform,. Considering the performance of Jetson Nano, we migrate some models to Jetson Nano for image recognition testing of pests and diseases.

In order to test the performance limit of the model, we run the Jetson Nano at full load and test 10892 images to calculate the accuracy and model inference speed. Table 5 shows the specific results of the test.

**Table 5** Comparisons of the performance of different models run on Jetson Nano. Inference speed is measured

by the number of images(224×224×3) the model infers per second, Underlined parameters represent the best parameters for that column.

| Model | Accuracy(%) | Inference speed |
| --- | --- | --- |
| **MobileNetV3-small+CA (ours)** | <u>98.31</u> | <u>272</u> |
| shufflenet_v2_x0_5 | 94.27 | 194 |
| shufflenet_v2_x1_0 | 97.53 | 222 |
| MobileNetV3-small | 95.83 | 253 |

On embedded devices, MobileNetV3-small+CA performs the best, with the best inference speed and inference accuracy, even surpassing the performance on high-performance computers. It shows that the improved model proposed in this paper is more suitable for edge computing scenarios with limited computing resources.

## 5 Conclusion

This paper proposes an improved MobileNetV3 lightweight crop pest identification method. It is used to solve the problems of difficult deployment and poor identification quality of pest identification models in agricultural production activities. The model has excellent performance in parameters, FLOPs, model size, and recognition accuracy. And the improvement of MobileNetV3-small performs well on embedded devices with limited computing resources, with fast inference speed and an accuracy rate of 98.31%.

The next step will be to study the identification of pests and diseases in complex scenarios to achieve real application value.